\documentclass[letterpaper]{article} 
\usepackage{aaai24}  
\usepackage{times}  
\usepackage{helvet}  
\usepackage{courier}  
\usepackage[hyphens]{url}  
\usepackage{graphicx} 
\usepackage{natbib}  
\usepackage{caption} 
\usepackage{algorithm}
\usepackage{algorithmic}

\usepackage{newfloat}
\usepackage{listings}
\DeclareCaptionStyle{ruled}{labelfont=normalfont,labelsep=colon,strut=off} 
\floatstyle{ruled}
\newfloat{listing}{tb}{lst}{}
\floatname{listing}{Listing}
\nocopyright

\title{Nested-TNT: Hierarchical Vision Transformers with Multi-Scale Feature Processing}

\author {
    Yuang Liu\textsuperscript{1,*},
    Zhiheng Qiu\textsuperscript{1,*},
    Xiaokai Qin\textsuperscript{1,*,\dag}
}

\affiliations{
    \textsuperscript{1} Equal contribution
 \\
    \textsuperscript{*} Nanyang Technological University, Singapore\\
    \textsuperscript{\dag} Corresponding author\\email: qi0002ai@e.ntu.edu.sg
}

\begin{document}

\maketitle

\begin{abstract}
Transformer has been applied in the field of computer vision due to its excellent performance in natural language processing, surpassing traditional convolutional neural networks and achieving new state-of-the-art. ViT divides an image into several local patches, known as "visual sentences". However, the information contained in the image is vast and complex, and focusing only on the features at the "visual sentence" level is not enough. The features between local patches should also be taken into consideration. In order to achieve further improvement, the TNT model is proposed, whose algorithm further divides the image into smaller patches, namely "visual words," achieving more accurate results. The core of Transformer is the Multi-Head Attention mechanism, and traditional attention mechanisms ignore interactions across different attention heads. In order to reduce redundancy and improve utilization, we introduce the nested algorithm and apply the Nested-TNT to image classification tasks. The experiment confirms that the proposed model has achieved better classification performance over ViT and TNT, exceeding 2.25\%, 1.1\% on dataset CIFAR10 and 2.78\%, 0.25\% on dataset FLOWERS102 respectively.
\end{abstract}

\section{Introduction}
The birth of Transformer \cite{vaswani2017attention} opens up the era of large models and accelerates the development of the field of natural language processing. The proposal of Transformer solves the drawbacks of traditional convolutional neural networks (CNNs) \cite{726791} that are weak in capturing global features, and improves the generalisation ability of the model. Compared with traditional recurrent neural networks, Transformer allows parallel computation, which improves computational efficiency. The core of the Transformer network is the self-attention mechanism, which focuses on the dependency of each element on the global, i.e., it enables the model to pay attention to the contextual information and allows bidirectional inference.

Inspired by Transformer's great success in the field of natural language processing, Vision Transformer (ViT) \cite{DBLP:journals/corr/abs-2010-11929} was born. As the name suggests, ViT transfers the idea of Transformer to the field of computer vision by segmenting an image into "visual sentences" to be fed into the Transformer network. ViT is widely used in the fields of image classification, image segmentation, target detection and image generation. For example, in the field of autonomous driving \cite{ando2023rangevit}, ViT has been applied to detect obstacles and traffic markers on the road. ViT has also been applied to image restoration and resolution enhancement \cite{lu2022transformer}, i.e., resolving a high-resolution version of a low-pixel image. ViT is also commonly used for target detection and target recognition in the medical field \cite{al2023vision}, biometrics field \cite{sun2022part} and video processing field \cite{arnab2021vivit}.

As mentioned earlier, ViT only splits images to the visual sentence level, which is still a coarse-grained approach. Therefore, ViT has limited accuracy in processing images, which can lead to loss of detailed features. In order to solve this problem and retain as much information as possible in the image, Transformer iN Transformer(TNT) \cite{DBLP:journals/corr/abs-2103-00112} was proposed. In TNT, image patches are further divided into "visual words", which provide a fine-grained approach to feature extraction.

On the other hand, the core of the Transformer architecture is a Multi-Head Attention (MHA) mechanism where each head is used to capture image features. In order to capture as many features as possible, a number of complex attention paradigms have been born under the existing TNT framework. In order to improve the efficiency of parameter utilization to reduce the model complexity, Nested Vision Transformer (Nested ViT) 
 \cite{jiquanpengandli2023improving} is proposed. Nested ViT maximizes the efficiency of parameter utilization by connecting the attention logistics of neighbouring layers of Transformers so that each head focuses on complementary information.

TNT and Nested ViT optimize some of the shortcomings of ViT, respectively. In order to combine the advantages of both we have designed a new method capable of simultaneously improving task accuracy and reducing parameter redundancy, called Nested-TNT. Nested-TNT refers to the idea of TNT and adds a layer of fine-grained Transformer Block on the basis of ViT to complete the extraction of image features in more details. In the original coarse-grained Transformer Block, the connection between different layers is established by MHA mechanism, which achieves the purpose of improving the efficiency of parameter utilization and reducing redundancy. We apply Nested-TNT to an image classification task and demonstrate its effectiveness experimentally.

\section{Related Works}
Just like a transformer, Vision Transformer (ViT) divides the input image into equally-sized patches, such as 16x16 pixels, which are analogous to "sequences" in a text sequence. Each patch is then transformed into a vector of a specific dimension through a linear projection, effectively creating an embedding for each image patch. At the start of the sequence, a "classification token" is introduced, and the ensuing sequence of vectors is then processed through a typical Transformer encoder. However, since the Transformer architecture inherently lacks the capability to process sequential data, positional embeddings are added to these patch embeddings. This step is crucial because it injects information into the model about the location of each patch in the image. Following this, the combined embeddings (patch and positional) are fed into several layers of the Transformer encoder. Within each layer, a multi-head self-attention mechanism and a multi-layer perceptron are employed. The output from the final Transformer encoder layer represents the aggregated output of the entire ViT model. It outperforms the convolutional neural networks (CNNs) on downstream tasks.

There is a lot of work dedicated to improving the performance and efficiency of ViT. For example, one team combined Convolutions and Vision Transformers to generate a new network CvT \cite{wu2021cvt}, which improves the performance compared to CNN-based models (e.g. ResNet) and prior Transformer-based models (e.g. ViT, DeiT) while utilising fewer FLOPS and parameters. Another example is LeViT \cite{graham2021levit}, which provides a better trade-off between accuracy and inference time, especially for small and medium-sized architectures. There also a team proposed the design of the Multiscale Visual Transformer (MViT) \cite{fan2021multiscale}, which introduces a multiscale feature hierarchy to the transformer. It uses the Multiple Heads Pooling Attention (MHPA) mechanism to allow flexible resolution modelling in the converter block. There are more relevant studies, such as Mvitv2 \cite{li2022mvitv2}, YOLOS \cite{fang2021you}, Pyramid vision transformer \cite{wang2021pyramid}, Reversible vision transformers \cite{mangalam2022reversible}, Scaling vision transformers \cite{zhai2022scaling}, Segmenter \cite{strudel2021segmenter} etc., which will not be expanded in the space limit.

Besides, among related studies, Transformer iN Transformer and Nested Vision Transformer caught our attention.

In ViT, the image is divided into local patches of 16x16 pixels, this coarse-grained division ignores more features and details. In order to obtain more accurate results, the Transformer iN Transformer (TNT) architecture is proposed. TNT further divides each local patch into smaller patches (4x4 pixels), i.e., each visual sentence is divided into visual words. Visual sentence inside the visual words, and then performs feature extraction and training in the next dimension at the visual sentence level. The algorithm achieves results beyond ViT.

The Multi-Head Attention (MHA) mechanism is the basis of the Transformer and Vision Transformer. As the network structure continues to be reconstructed and overlaid, the number of head parameters becomes increasingly large. However, it is not always necessary to have so many header parameters. Peng's team \cite{jiquanpengandli2023improving} noticed that if it is possible to make each head pay attention to as much complementary information as possible, we can maximize the efficiency of parameter utilization and reduce the redundancy. Based on the "explaining-away" effect, they found a way to make the features of each head complementary called Nested Vision Transformer (Nested ViT). They established connections between layers of the Transformer to enable the network to capture both detailed and global features, and experimentally demonstrated that this approach effectively reduces parameter redundancy.

\begin{figure*}[ht]
    \centering
    \includegraphics[width=0.85\linewidth]{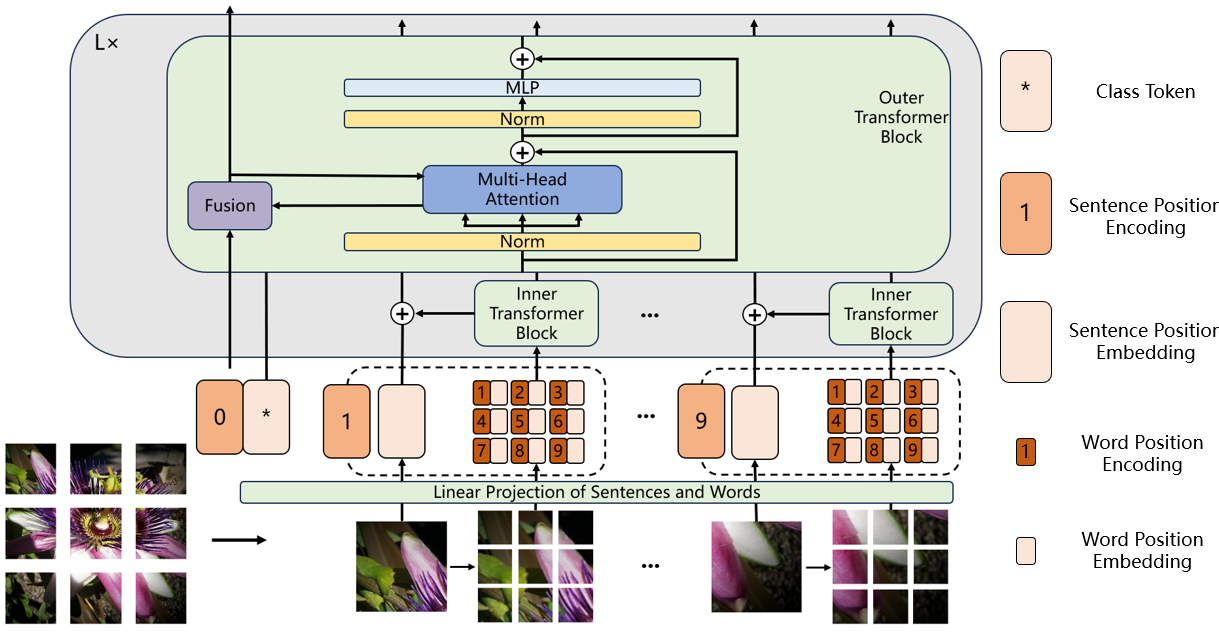}
    \caption{Nested Multi-head Attention TNT Structure}
    \label{fig:structure}
\end{figure*}

\section{Methodology}
Firstly, we will provide a brief outline of the structure underlying multi-head attention. Multi-Head Attention algorithm sets independent queries, keys and values matrices for every single head and executes the attention procedure simultaneously across these heads. Subsequently, it combines the output values from each head through concatenation, followed by linear projection to generate the ultimate output. If we use $Q_{i}$, $K_{i}$, $V_{i}$ and $H_{i}$ to represent queries, keys, values and the output of a single head, then the calculation process is shown in the following formula:

$$
 H_{i} = softmax\left(\frac{Q_{i} K_{i}^{T}}{\sqrt{d}}\right) V   (1)
$$

Through linear projection, we can reach the final output of multi-head attention: $\sum_{i=1}^{h}H_{i}W_{i}^{o}$, where $W_{i}^{o}$ are learnable weights.

An overview of our model is depicted in Figure \ref{fig:structure}. Our usage of transformer is partly based on TNT \cite{DBLP:journals/corr/abs-2103-00112}. In total, there are L layers. Each layer comprises multiple inner transformers, treating words as basic units, along with an outer transformer focusing on sentences as basic units. Standard learnable 1D position encodings are applied both in word embeddings and sentence embeddings to keep spatial information. Sentence position encoding can maintain the global spatial information, while word position encoding is used for preserving the local relative position, and the word position encodings are shared across sentences.

To be more specific, the input picture is initially divided into n patches as ViT \cite{DBLP:journals/corr/abs-2010-11929} does and each patch is regarded as a visual sentence. Then each visual sentence is split into a sequence of visual words. After word embedding, the vectors are sent to inner transformer blocks, whose purpose is to learn the relationships among visual words. For an embedded visual sentence $Y_{l}^{i} =[y_{l}^{i,1},y_{l}^{i,2},...,y_{l}^{i,m}] $, $l = 1,2,...,L$ is the index of the block, where $i$ is the index of visual sentence and there are $m$ visual words in the sentence. The calculation process of the inner transformer can be expressed by the following formula, where $MSA()$ is the multi-head attention process, $LN()$ is the layer normalization process and $MLP()$ stands for the layer applied between self-attention layers for feature transformation and non-linearity.    
$${Y^{'}}_{l}^{i} = Y_{l-1}^{i} + MSA(LN(Y_{l-1}^{i})) (2)$$
$$Y_{l}^{i} = {Y^{'}}_{l}^{i} + MLP(LN(Y_{l-1}^{i})) (3)$$
After transformation of inner transformers, we have the expression for all word embeddings:$\gamma_{l} =[Y_{l}^{1},Y_{l}^{2},...,Y_{l}^{n}] $.    

At sentence level, we do sentence embedding and the procedure can be depicted as $\beta_{0} =[Z_{class},Z_{0}^{1},...,Y_{0}^{n}]$ where $Z_{class}$ is the class token. Before entering outer transformer, the sentence embedding should be augmented by word-level features learned by inner transformers:
$$Z_{l-1}^{i} = FC(Vec(Y_{l}^{i})) (4)$$
$FC()$ stands for fully connected layer and $Vec(·)$ is the vectorization operation. Then we use outer transformer block to deal with the sentence embeddings:
$${Z^{'}}_{l} = Z_{l-1} + MSA^{'}(LN(Z_{l-1})) (5)$$
$$Z_{l} = {Z^{'}}_{l} + MLP(LN(Z_{l-1})) (6)$$
where $MSA^{'}()$ stands for nested multi-head attention mechanism. The mechanism creates a new data path so that the attention logits of adjacent layers' outer transformer blocks can communicate directly. The addition of this mechanism was inspired by Peng's team\cite{jiquanpengandli2023improving}, who also pointed out the impact of the explaining-away effects when stacking transformer blocks. Specifically, when an attention head focuses on parts of the input that are highly relevant to the output, it prevents other heads from doing the same things. On the contrary, it encourages them to focus on supplementary information. This means that proper connections between the heads will help reduce the redundancy and improve the efficiency of model parameters. Multi-head attention mechanism provides communication methods for different attention heads within each layer, but the connection between heads from different layers is still weak. The data path uses a multi-layer perceptron layer to fuse the attention logits from adjacent layers: 
$$A_{outer}^{l} = softmax(z^{l}_{outer} + MLP(z^{l}_{outer},z^{l-1}_{outer})) (7)$$
where $z^{l}_{outer}$ stands for the attention logits of the out transformer of layer l and $A_{outer}^{l}$ is the attention weights of the out transformer of layer l. During training, softmax cross-entropy loss function and gradient descent optimizer are applied to update the parameters. 

To sum up, according to the structure shown in Figure \ref{fig:structure} and the formulas listed above, the inputs and outputs of Nested TNT can be formulated as 
$$\gamma_{l},\beta_{l},z_{outer}^{l} = Nested\_TNT(\gamma_{l-1},\beta_{l-1},z_{outer}^{l-1}) (8)$$

\section{Experiments}
In this section, we conduct experiments on several visual datasets to evaluate the classification ability of the proposed Nested Multi-head Attentions TNT architecture. Moreover, two baseline models (ViT\cite{DBLP:journals/corr/abs-2010-11929}, TNT \cite{DBLP:journals/corr/abs-2103-00112}) are chosen as comparison targets and also adopted with the same experiments.
\subsection{Datasets}
CIFAR100, CIFAR10 \cite{krizhevsky2009learning}, are two image classification benchmarks including 50000 train and 10000 test images belonging to 100 and 10 classes respectively. The images included in these two datasets are in the same size, 32x32 with 3 color channels. All the classes have same number of images, which helps avoid bias in the training process. CIFAR100 is used as pre-train dataset while CIFAR10 is used as fine-tune dataset. 

In addition to CIFAR10, Flowers102 \cite{gogul2017flower} is the other fine-tune dataset, where some flower categories are very similar in appearance. This is a challenge for classification algorithms and requires models that can capture and learn subtle feature differences. 

Adopted data augmentation strategies are similar to those in Nested-Attention ViT \cite{jiquanpengandli2023improving} implementation, e.g., random crop, random flip, mixup and bicubic train-interpolation\cite{chen2023nonuniform}. The details of visual datasets are shown in Table \ref{Datasets}. 

\begin{table}[h]
\centering
\begin{tabular}{|c|c|c|c|c|}
\hline 
\textbf{Dataset}&\textbf{Type}&\textbf{Train Size}&\textbf{Test Size}&\textbf{Classes}\\
\hline
\textbf{Cifar100}&Pre-train&50000&10000&100\\
\hline
\textbf{Cifar10}&Fine-tune&50000&10000&10\\
\hline
\textbf{Flowers102}&Fine-tune&6149&2040&102\\
\hline
\end{tabular}
\caption{Details of Used Datasets.}
\label{Datasets}
\end{table}

\subsection{Experimental Settings}
The 3 models chosen in our experiment are ViT-S, TNT-S, and small size of the proposed architecture, which all process images at 224x224-pixels resolution. Moreover, the main settings for hyper-parameters in pre-training and fine-tuning are shown in tables \ref{Pre-Train Hyper-Parameters} and \ref{Fine-Tune Hyper-Parameters} respectively.

\begin{table}[ht]
\centering
\begin{tabular}{|c|c|c|}
\hline
\textbf{Epochs}&\textbf{Optimizer}&\textbf{Batch Size}\\
\hline
150&SGD&32\\
\hline
\textbf{Learning Rate}&\textbf{Learning Decay}&\textbf{Weight Decay}\\
\hline
5e-2&Cosine&1e-4\\
\hline
\textbf{Warmup epochs}&\textbf{Label Smooth}&\textbf{train-interpolation}\\
\hline
3&0.1&bicubic\\
\hline
\end{tabular}
\caption{Pre-Train Hyper-Parameters.}
\label{Pre-Train Hyper-Parameters}
\end{table}

\begin{table}[ht]
\centering
\begin{tabular}{|c|c|c|}
\hline
\textbf{Epochs}&\textbf{Optimizer}&\textbf{Batch Size}\\
\hline
100&SGD&32\\
\hline
\textbf{Learning Rate}&\textbf{Learning Decay}&\textbf{Weight Decay}\\
\hline
5e-2&Cosine&1e-4\\
\hline
\textbf{Warmup epochs}&\textbf{Label Smooth}&\textbf{train-interpolation}\\
\hline
3&0.1&bicubic\\
\hline
\end{tabular}
\caption{Fine-Tune Hyper-Parameters.}
\label{Fine-Tune Hyper-Parameters}
\end{table}

\subsection{Results}
The parameter complexity for three models is shown in Table \ref{Parameters}. Nested-TNT model has the highest parameter complexity, but only a little higher than that of TNT. And it is obviously more complex than that of VIT, which is one of the sacrifices our model makes for the following advantages.

\begin{table}[h]
\centering
\begin{tabular}{|c|c|c|c|}
\hline 
\textbf{Models}&ViT&TNT&Nested-TNT\\
\hline
\textbf{Params(M)}&21.67&23.41&23.42\\
\hline
\end{tabular}
\caption{Parameters for Three Models.}
\label{Parameters}
\end{table}

The results of classification accuracy are shown in Table \ref{Accuracy}. The accuracy values are all in top-1 accuracy form. The Nested-TNT outperforms the other two visual transformer models in terms of final accuracy, although it is reached at the expense of number of images processed per second. For CIFAR100 and CIFAR10, Nested-Attention TNT's better performance comes from the introduced Nested-Attention, which enhances links between global modules. Then, better links create better recognition.

For Flowers102, the higher accuracy means Nested-Attention TNT has better performance in fine-grained analysis. This comes from the added inner blocks in TNT, which helps captures localized details of an image. As a result, it gets a result that is similar to that of TNT and 3\% higher than that of ViT.

\begin{table}[ht]
\centering
\begin{tabular}{|c|c|c|c|c|}
\hline 
\textbf{Model}&Cifar100&CIFAR10&Flower&im/sec\\
\hline
\textbf{ViT}&66.98&91.28&90.49&522\\
\hline
\textbf{TNT}&69.63&92.43&93.02&190\\
\hline
\textbf{Nested-TNT}&\textbf{70.59}&\textbf{93.53}&\textbf{93.27}&137\\
\hline
\end{tabular}
\caption{Accuracy for Three Models.}
\label{Accuracy}
\end{table}

\section{Conclusion}
We propose a new Vision Transformer model, Nested-TNT, based on nested multi-head attention mechanism and principle of Transformer iN Transformer (TNT) model. The nested multi-head attention mechanism improves independence between different attention heads, while the TNT model cuts the image patches into smaller patches. Then, the experiment results show that Nested-TNT performs better on image classification task. It demonstrates its ability to enhance both the detailed and the global features at the same time, which satisfies our basic targets.

\section{Limitation and Future Work}
Although our proposed model has a better performance on the image classification accuracy, the image processing speed and parameter complexity are not so optimistic. Also, because of the time and computing limitation, the amount of pre-train is insufficient.

Thus, in the future, firstly, the extension of the Nested-TNT can be focused on the simplification of the algorithms and the optimization of connection layers. After that, we can apply more pre-train to get better performance. The pre-train dataset can be switched to a larger one, e.g., ImageNet. Furthermore, the experiments were limited to categorization tasks, which can be extended to Objective Detection and Semantic Segmentation fields. Ablation study can be taken into consideration to check each part's ability as well.

\bibliography{Nested-TNT}

\section{About Authors}
This research was jointly completed by three individuals.

\textbf{Yuang Liu}
has a bachelor degree in Electronics and Computer Science from University of Edinburgh. Now he is pursuing his Master's degree in Artificial Intelligence at Nanyang Technological University.

\textbf{Zhiheng Qiu}
has a Bachelor's Degree in Computer Science and Technology from Zhejiang University, and pursuing a Master's degree in Artificial Intelligence at Nanyang Technological University currently.

\textbf{Xiaokai Qin}
has a B.Sc. degree in Computer Science from University of Liverpool and B.Sc. degree in Information and Computing Science from Xi'an Jiaotong Liverpool University. Presently, he is pursuing his Master's degree in Artificial Intelligence at Nanyang Technological University, Singapore. 

\end{document}